\documentclass[conference]{IEEEtran}
\IEEEoverridecommandlockouts
\usepackage{stfloats}
\usepackage{makecell}
\usepackage{url}
\usepackage{cite}
\usepackage{amsmath,amssymb,amsfonts}
\usepackage{algorithmic}
\usepackage{graphicx}
\usepackage{textcomp}
\usepackage{xcolor}
\def\BibTeX{{\rm B\kern-.05em{\sc i\kern-.025em b}\kern-.08em
    T\kern-.1667em\lower.7ex\hbox{E}\kern-.125emX}}
\begin{document}

\title{MiLMo:Minority Multilingual Pre-trained Language Model\\
\thanks{Supported by National Nature Science Foundation (No. 61972436).}
}

\author{
    \IEEEauthorblockN{Junjie Deng$^{1,2,\&}$, Hanru Shi$^{1,2,\&}$, Xinhe Yu$^{1,2}$, Wugedele Bao$^{3}$,  Yuan Sun $^{1,2,*}$, Xiaobing Zhao$^{1,2,*}$}
    \IEEEauthorblockA{$^1$ Minzu University of China, China}
    \IEEEauthorblockA{$^2$ National Language Resource Monitoring \& Research Center Minority Languages Branch}
    \IEEEauthorblockA{$^3$ Hohhot Minzu College}
    *Corresponding author: Yuan Sun, Xiaobing Zhao\\
    $^\&$These authors contributed equally to this work and should be considered co-first authors
    \IEEEauthorblockA{junjie\_deng\_cn@163.com, 1834171972@qq.com, tracy.yuan.sun@gmail.com, nmzxb\_cn@163.com}
    }

\maketitle

\begin{abstract}
Pre-trained language models are trained on large-scale unsupervised data, and they can fine-turn the model only on small-scale labeled datasets, and achieve good results. Multilingual pre-trained language models can be trained on multiple languages, and the model can understand multiple languages at the same time. At present, the search on pre-trained models mainly focuses on rich resources, while there is relatively little research on low-resource languages such as minority languages, and the public multilingual pre-trained language model can not work well for minority languages. Therefore, this paper constructs a multilingual pre-trained model named MiLMo that performs better on minority language tasks, including Mongolian, Tibetan, Uyghur, Kazakh and Korean. To solve the problem of scarcity of datasets on minority languages and verify the effectiveness of the MiLMo model, this paper constructs a minority multilingual text classification dataset named MiTC, and trains a word2vec model for each language. By comparing the word2vec model and the pre-trained model in the text classification task, this paper provides an optimal scheme for the downstream task research of minority languages. The final experimental results show that the performance of the pre-trained model is better than that of the word2vec model, and it has achieved the best results in minority multilingual text classification. The multilingual pre-trained model MiLMo, multilingual word2vec model and multilingual text classification dataset MiTC are published on \url{http://milmo.cmli-nlp.com/}.
\end{abstract}


\section{Introduction}
With the development of deep learning, various neural networks are widely used in the downstream tasks of natural language processing and have achieved good performance. These downstream tasks usually rely on a large-scale of labeled training datasets, but labeled datasets often require significant human and material resources. The emergence of pre-trained language model~\cite{1,2,3,4,5} has solved this problem well. The pre-trained model is trained on large-scale unsupervised data to obtain a general model. In the downstream tasks, the model can achieve good performance only by fine-tuning the small-scale labeled data, which is crucial for the research on low-resource languages.

BERT~\cite{1} is the most influential model among various pre-trained language models, which has achieved the best results in a variety of downstream tasks. However, there are still some problems in BERT, and a large number of BERT variant pre-trained language models~\cite{3,14,15} have emerged to solve these problems, but these researches are mainly focused on large-scale resources such as English, and there are still few researches on low resources. Moreover, these pre-trained models can only be trained in a single language, and there is no shared knowledge between different language models. To solve these problems, multilingual pre-trained language models~\cite{16,17,18,37} come into being, which can process multiple languages at the same time. The existing multilingual pre-trained model is trained on an unlabeled multilingual corpus, which can project multiple languages into the same semantic space, and has the ability of cross-lingual transfer. It can conduct zero-shot learning.

At present, the pre-trained language model has been well developed in large-scale languages. However, for minority languages, since it is difficult to obtain corpus resources and there are relatively few related studies, various publicly available multilingual pre-trained models do not work well on minority languages, which seriously affects the construction of minority language informatization. Moreover, the existing multilingual pre-trained models only include a few minority languages. Although cross-lingual transfer can be applied to minority languages, the effect is not ideal. For example, the F1 of mBERT~\cite{16} on Tibetan News Classification
Corpus (TNCC)~\cite{26} is 5.5$\%$~\cite{34}, and the F1 of XLM-R-base on TNCC is 21.1$\%$~\cite{34}. To further promote the development of natural language processing tasks in minority languages, this paper has fully collected and sorted out relevant documents from the Internet, relevant books, the National People's Congress (NPC) and the Chinese Political Consultative Conference the National People's Congress (CPPCC), and government work reports, and trains a multilingual pre-trained model named MiLMo on these data. The main contributions of this paper are as follows:

\begin{itemize}
\item This paper constructs a pre-trained model MiLMo containing five minority languages, including Mongolian, Tibetan, Uygur, Kazakh and Korean, to provide support for various downstream tasks of minority languages.
\item This paper trains a word2vec representation for five languages, including Mongolian, Tibetan, Uygur, Kazakh and Korean. Comparing the word2vec representation and the pre-trained model in the downstream task of text classification, this paper provides the best scheme for the research of downstream task of minority languages. The experimental results show that MiLMo model outperforms the word2vec representation.
\item  To solve the problem of scarcity of minority language datasets, this paper constructs a classification dataset MiTC containing five languages, including Mongolian, Tibetan, Uyghur, Kazakh and Korean, and publishes the word2vec representation, multilingual pre-trained model MiLMo and multilingual classification dataset MiTC on \url{http://milmo.cmli-nlp.com/}.
\end{itemize}

\section{Related Work}

Word representation can convert words in natural language into vectors which is of great significance to natural language tasks. Early word vectors can capture the semantics of words, but they are context independent and cannot solve the problem of polysemy~\cite{6,7,8}. To solve the above problems, relevant researchers study context-sensitive word representation. ELMo~\cite{9} is the first model to apply context-sensitive word representation successfully. It uses language model to learn the word vector representations in a large-scale unsupervised database, and then the word representations of the network layers corresponding to the words are extracted from the pre-trained network as new features to be added to the downstream task, and the word vector representations can be obtained using the contextual information of the data. However, this model splices two unidirection LSTMs, and the ability of feature extraction and fusion is weak.

In 2017, Google proposes the Transformer block~\cite{10} in the machine translation task, which is a new encoder-decoder architecture that uses the attention mechanism to encode each location and can parallelize processing. Transformer solves the problem that CNN needs a lot of storage resources to remember the whole sequence and the inability to parallelize processing. On this basis, OpenAI proposes the GPT, which uses the Transformer block with 12 layers. GPT is first pre-trained on an unlabeled dataset to obtain a language model, and then deal with various downstream tasks by fine-tuning. GPT-1~\cite{4} can achieve good results after fine-tuned, but does not work well on the tasks without fine-tuning. To train a word vector model with stronger generalization ability, GPT-2~\cite{12} uses more network parameters and larger datasets. GPT-2 considers supervised task as the sub-task of the language model. It verifies that word vector models trained with massive data and parameters can be directly transferred to other tasks without fine-tuning on labeled data. To further improve the effect of unsupervised learning, GPT-3~\cite{13} further increases training data and parameters, and achieves better results in a variety of downstream tasks.

However, GPT is a unidirection language model. BERT proposes to use MLM to pre-train to generate deep bidirection language representation. After BERT, various BERT variants have emerged, such as ALBERT~\cite{14}, SpanBERT~\cite{15} and RoBERta~\cite{16}. These models have achieved better results. However, these studies are mainly monolingual and focus on large-scale languages, such as English. There is still little research on low resources. To solve this problem, multilingual pre-trained models~\cite{16,17,18,19} begin to emerge. Facebook AI Research proposes the XLM\cite{17}, which uses Byte Pair Encoding (BPE) to preprocess training data, and expands the shared vocabulary between different languages by dividing text into sub-words. They propose three pre-training tasks, including CLM, MLM, and TLM, which have achieved good results in cross-lingual classification tasks. After that, Facebook AI Research put forward XLM-R. This model increases the number of languages in the training dataset on the basis of XLM and RoBERTa, and upsamples low-resource languages in the process of vocabulary construction and training to generate a larger shared vocabulary and improve the ability of the model. Liu et al. propose mBERT~\cite{16}, they select largest 104 languages in Wikipedia as the training data, they also train the model in MLM and NSP tasks. They use the same model and weight to process all target languages, and make the model have cross-lingual transfer capabilities by using the shared parameters. The above cross-lingual models have achieved great success in large-scale languages, such as English. However, due to the scarcity of minority language corpus and the complexity of grammar rules, the above multilingual models can not deal with them well. Therefore, the research on various downstream tasks in minority languages is still at an early stage, which seriously hinders the development of minority natural language processing. To solve the above problems, The HIT·iFLYEK Language Cognitive Computing Lab releases the minority language pre-trained model CINO~\cite{34}, which provides the ability to understand Tibetan, Mongolian, Uyghur, Kazakh, Korean, Zhuang, Cantonese, Chinese and dialects. To further promote the development of ethnic minority language pre-trained model, this paper trains a multilingual pre-trained model for ethnic minority languages. The experimental results show that our model can effectively promote the research on the downstream tasks of ethnic minority languages.

\begin{table}[h]
\begin{center}
\setlength{\tabcolsep}{10mm}{
\begin{tabular}{c|c}
\hline \bf Languages & \bf The Amount of Data \\ \hline
Mongolian       & 788MB\\
Tibetan                   & 1.5GB\\
Uyghur               & 397MB\\
Kazakh   & 620MB\\
Korean                 & 994MB\\
\hline
\end{tabular}}
\end{center}
\renewcommand{\tablename}{Table}
\caption{\label{font-table} The amount of data for each minority language}
\end{table}

\begin{table*}[h]
\centering
\begin{tabular}{cc}
\hline \bf Languages & \bf Segmentation \\ \hline
Mongolian & \makecell[c]{\\\includegraphics[width=0.4\textwidth]{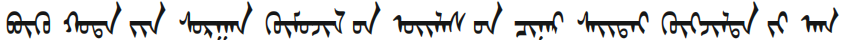}\\(High-quality development of education in the city.)}\\

Tibetan\_syll & \makecell[c]{\\\includegraphics[width=0.7\textwidth]{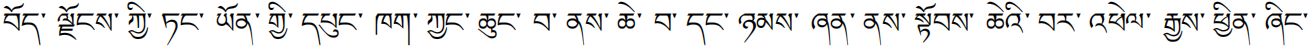} \\(The ranks of Tibetan party members have also changed from small to large, from weak to strong.)}\\

Tibetan\_word & \makecell[c]{\\\includegraphics[width=0.7\textwidth]{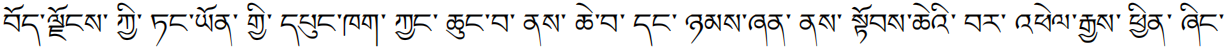}\\(The ranks of Tibetan party members have also changed from small to large, from weak to strong.)} \\

Uyghur & \makecell[c]{\\\includegraphics[width=0.6\textwidth]{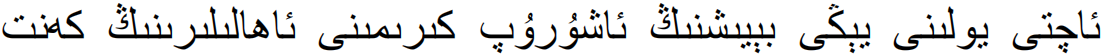}\\(Open up new ways for villagers to increase income and become rich.)}\\

Kazakh & \makecell[c]{\\\includegraphics[width=0.6\textwidth]{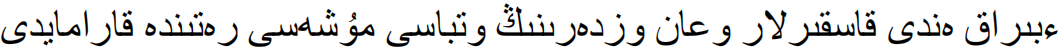}\\(Now they don't see him as part of the family anymore.)}\\

Korean & \makecell[c]{\\\includegraphics[width=0.4\textwidth]{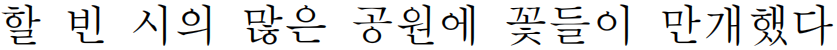}\\(Many parks in Harbin are in full bloom.)} \\

\hline
\end{tabular}

\caption{\label{font-table} The results of word segmentation in each minority language}
\end{table*}

\begin{figure*}[h] 
\setlength{\abovecaptionskip}{1pt}
\setlength{\belowcaptionskip}{2pt}
    \centering 
    \includegraphics[width=1.0\textwidth]{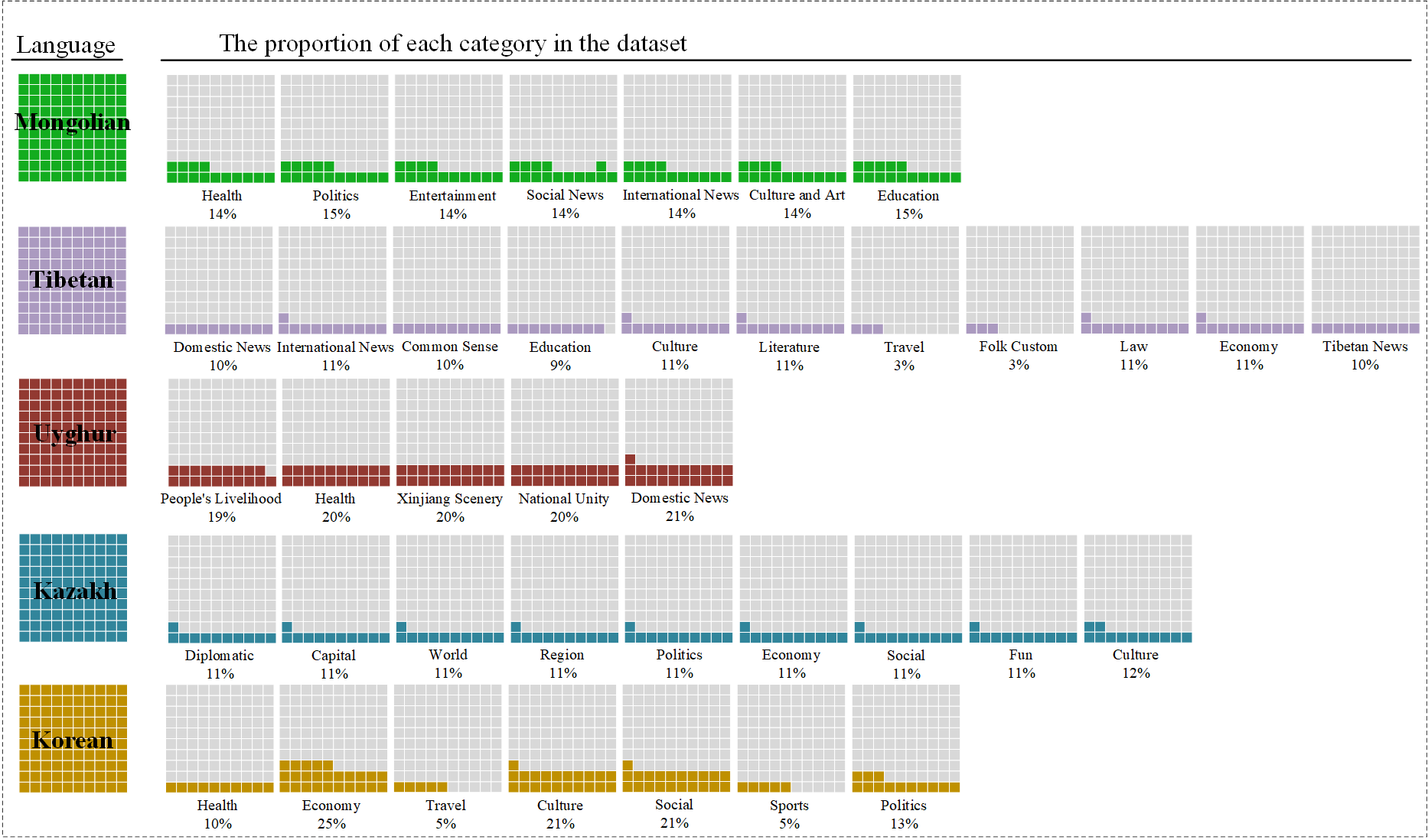}
    \renewcommand{\figurename}{Figure}
    \caption{Distribution of text types in the MiTC} 
    \label{wolf} 
\end{figure*}

\section{Model Details}
\subsection{Word2vec's Data Preprocessing}
This paper obtains the training data of five minority languages, including Mongolian, Tibetan, Uyghur, Kazakh and Korean from the network, relevant books, the NPC and CPPCC sessions, government work reports and other relevant documents. And we delete the non-text information such as pictures, links and symbols, and discard the articles with text length less than 20 to clean the data. The final data information is shown in Table I.

Before training the model, we need to segment the data. The above five minority languages are all alphabetic writings. Mongolian is spelled with words from top to bottom and left to right. Mongolian words are separated by spaces~\cite{22}, so this paper directly uses spaces for word separation. The smallest unit of a Tibetan word is a syllable, and a syllable contains one or up to seven characters. The syllables contain rich semantic information. Therefore, this paper segments Tibetan sentences at syllable level and word level respectively~\cite{35}. 
The morphological structure of Uyghur words is complex. There are 32 letters in modern Uyghur, and each word is spelled by letters. The end of the word realizes its grammatical function by pasting different affixes. Therefore, the same word root can evolve into different word morphologies without great differences in word meanings. Uyghur is written from right to left, and words are separated by spaces. Each Uyghur word can be used as a feature item~\cite{24}. Therefore, this paper uses spaces to segment Uyghur words. In Kazakh, words are also separated by spaces~\cite{23}. In Korean, the space cannot be used as a direct word segmentation mark. Morpheme is the smallest linguistic unit with semantics, so it is necessary to segment sentences into morphemes. In this paper, the Korean processing toolkit KoNLPy~\cite{21} proposed by Park et al. is used to segment the Korean corpus, and the morphemes obtained are used as input features. The word segmentation results of the above five languages are shown in Table II. This paper uses skip-gram model to train word representation with 300 dimension.

\subsection{Pre-trained Model MiLMo}
\subsubsection{MiLMo's Design}
XLM is a cross-lingual pre-trained model proposed by Facebook AI Research. This model can be trained in multiple languages, allowing the model to learn more cross-lingual information, so that the information learned from other languages can be applied to low-resource languages. XLM proposes three pre-training tasks, including Causal Language Modeling (CLM), Masked Language Modeling (MLM) and Translation Language Modeling (TLM).
CLM is a Transformer language model, which can predict the probability of the next word of a given sentence$P(w_{t}|w_{1},...,w_{t-1},\Theta)$. MLM is a masking task in which the model masks the tokens in the input sentence with a certain probability and then predicts the masked tokens. TLM is an extension of MLM, which splices parallel corpora from different languages as the input of the model, then masks some of these tokens according to a certain probability, and then predicts these tokens. CLM and MLM only need unsupervised training on monolingual datasets, and TLM needs supervised learning using parallel corpus. This paper uses MLM model. The input of the model is a sentence with 256 tokens. During training, part of the tokens are masked with a probability of 15$\%$. For each masked token, 80$\%$ uses [mask] instead, 10$\%$ randomly selects a token from the vocabulary, and 10$\%$ remains unchanged. The model parameters are shown in Table III. The MiLMo model is trained using the 12 layer Transformer blocks.

\begin{table}[h]
\begin{center}
\setlength{\tabcolsep}{10mm}{
\begin{tabular}{c|c}
\hline \bf Parameter & \bf Value \\ \hline
emb\_dim & 2048\\
n\_layers & 12 \\
n\_heads &  8\\
dropout & 0.1 \\
n\_langs & 5\\
max\_len & 256 \\
vocab\_size & 70,000\\
\hline
\end{tabular}}
\end{center}
\renewcommand{\tablename}{Table}
\caption{\label{font-table} The training parameters of the MiLMo}
\end{table}
\subsubsection{Shared sub-word vocabulary}
BPE~\cite{25} is a data compression algorithm. By iteratively merging character pairs with high frequency, variable length sub-words can be generated in a fixed size vocabulary. The process of building a vocabulary is as follows: 

\begin{itemize}
\item Divideing the words in a sentence into individual characters and uses all the characters to build an initial vocabulary.
\item Counting the frequencies of adjacent sub-word pairs within words in the corpus. 
\item Selecting the sub-word pairs with the highest frequency, merge them into new sub-words, and add new sub-words to the sub-word vocabulary. 
\item Deleting the sub-words that no longer exist in the corpus from the vocabulary. 
\end{itemize}

\begin{table}[h]
\begin{center}
\setlength{\tabcolsep}{8mm}{
\begin{tabular}{c|cc}
\hline \bf Language & \bf MiTC  & \bf WCM \\ \hline
Mongolian & 1,747 & 2,973 \\
Tibetan &4,926  & 1,110\\
Uyghur & 1,304  & 300\\
Kazakh & 35,826 & 6,258\\
Korean & 38,859 & 6,558\\
Total   &  82,662   &17,199   \\
\hline
\end{tabular}}
\end{center}
\renewcommand{\tablename}{Table}
\caption{ Language scale statistics in WCM and MiTC}
\end{table}

\begin{table*}[h]
\begin{center}
\begin{tabular}{ccccccc}
\hline 
\bf Model     & \bf Mongolian          & \bf Tibetan\_syll  & \bf Tibetan\_word     & \bf Uyghur      & \bf Kazakh        & \bf Korean         \\ \hline
TextCNN	     &	35.12\%             &\textbf{39.60\%}    &	32.85\%           &34.59\% &	27.52\%            &	\textbf{53.17\%}  \\
TextRNN	     &	28.77\%             &26.26\%             &	31.48\%           &\textbf{42.19\%} &	17.95\%            &	43.22\%           \\
TextRNN\_Att &	22.17\%             &	24.04\%           &	17.20\%           &28.93\% &	20.21\%            &	38.88\%           \\
TextRCNN     &	27.81\%             &	28.92\%           &	26.37\%           &38.23\% &	24.40\%            &	43.41\%           \\
FastText     &	17.32\%             &	11.23\%           &	16.02\%           &26.34\% &	11.09\%            &	19.42\%           \\
DPCNN        &	\textbf{49.15\% }   &	34.79\%           &\textbf{	34.51\% } &32.15\% &	\textbf{30.13\%}   &	52.85\%           \\
Transformer	 &	24.13\%            &26.53\%              &	18.67\%           &34.90\% &	10.88\%            &	33.63\%           \\

\hline
\end{tabular}
\end{center}
\renewcommand{\tablename}{Table}
\caption{\label{font-table} Text classification results of word2vec on MiTC}
\end{table*}

\begin{table*}[h]
\begin{center}
\setlength{\tabcolsep}{5mm}{
\begin{tabular}{ccccccc}
\hline 
\bf Model     & \bf Mongolian            & \bf Tibetan     & \bf Uyghur      & \bf Kazakh        & \bf Korean         \\ \hline
word2vec\_best&	49.15\%          &	34.51\%        & 38.23\%           &	30.13\%           &	53.17\%              \\
MiLMo-base   &	\textbf{81.48\%}  &	\textbf{76.44\%}&\textbf{74.13\% }   &	\textbf{71.34\%}   &	\textbf{85.98\%}     \\

\hline
\end{tabular}}
\end{center}
\renewcommand{\tablename}{Table}
\caption{\label{font-table} Comparison of text classification results between word2vec and MiLMo on MiTC}
\end{table*}

From the process of building the vocabulary, BPE is generally applicable to character formal languages involving prefixes and suffixes. The five languages in this paper are all alphabetic writings with stickiness phenomenon.  Grammar functions can be realized by pasting different suffixes in the front, middle and back of the root words. Therefore, BPE can be used to preprocess the five minority languages and build a shared vocabulary, so that it can effectively improve the efficiency of word segmentation and simplify the vocabulary. Before training the model, this paper splits the training corpus into training set, validation set, and test set in with ratio 8:1:1, and then the BPE is used to preprocess the training corpus.

To cover all the corpus to the greatest extent, this paper combines all the training sets of five languages, and uses BPE to build the vocabulary. The final vocabulary has 70,000 sub-words and covers 99.95$\%$ of the training sets. Then we use the trained BPE model and the constructed vocabulary to preprocess the training data of five minority languages.

\section{Experiments}
\subsection{Dataset}
Due to the difficulty in acquiring minority languages and the complexity of grammar rules, at present, the only publicly available dataset on minority languages is Wiki-Chinese-Minority (WCM)~\cite{34} , a minority language classification task dataset from Harbin Institute of Technology. This dataset is based on the Wikipedia corpus of minority languages and its classification system labels, including Mongolian, Tibetan, Uyghur, Cantonese, Korean, Kazakh and Chinese. It covers ten categories of art, geography, history, nature, natural science, characters, technology, education, economy and health. The total number of samples of five languages in WCM is 17,199, including Mongolian, Tibetan, Uyghur, Kazakh and Korean. While training the pre-trained model of minority languages, this paper constructs a multilingual text classification dataset MiTC for various languages, which contains 82,662 samples. The data of each language in WCM and MiTC is shown in Table IV.

From Table IV, we can see that the MiTC dataset is rich in all five languages, and the number of samples in Tibetan, Uyghur, Kazakh and Korean is much larger than the number of samples in WCM. After analyzing WCM, we find that the number of samples in the same language is not balanced. For example, the Tibetan contains eight categories, of which the number of "education" categories accounts for a small proportion. The Uyghur contains six categories. The total sample contains 300 pieces of data, but the "geography" category has 256 samples. The unbalanced distribution of samples leads to low accuracy of the model, and the performance of the model cannot be evaluated. To better evaluate the performance of the model, this paper performs a data balancing process on the MiTC dataset by keeping the data volume of each category relatively balanced under the same language. The distribution of the categorical dataset of each language in the MiTC dataset is shown in Figure I.

\subsection{Classification based on Word2vec}
This paper trains word2vec representation for five minority languages. Word2vec solves the problems of "dimension disaster" and document vector sparsity. It transforms each word into a low-dimensional real vector based on the contextual semantic information of the document. The more similar the meaning of words is, the more similar they are in the word vector space. This paper uses skip-gram to train 300-dimensional word representation, and uses the word  representation for text classification tasks. We used TextCNN, TextRNN, TextRNN Attention, TextRCNN, FastText, DPCNN and Transformer models to conduct classification experiments on MiTC. TextRNN Attention is a bidirectional LSTM network based on the Attention mechanism. The F1 of classification results is shown in Table V.

\begin{table*}[h]
\begin{center}
\begin{tabular}{ccccccc}
\hline 
\bf Model     & \bf Mongolian          & \bf Tibetan\_syll  & \bf Tibetan\_word     & \bf Uyghur      & \bf Kazakh        & \bf Korean         \\ \hline
TextCNN	      &	55.52\%             &\textbf{30.63\%}  &	43.01\%           &	69.44\%          &	42.08\%             &	\textbf{30.44\%}   \\
TextRNN	      &	55.99\%             &17.65\%           &	37.61\%           &	69.44\%          &	39.02\%             &	18.01\%            \\
TextRNN\_Att  &	31.85\%             &25.62\%           & 26.37\%            &	69.44\%          &	25.18\%             &	7.41\%             \\
TextRCNN	  &	55.53\%             &19.07\%           &	46.98\%           &	69.44\%          &	32.35\%             &	17.53\%            \\
FastText	  &	31.85\%             &17.34\%           &	27.23\%           &	69.44\%          &	10.56\%             &	16.77\%           \\
DPCNN	      &	\textbf{56.69\%}    &30.01\%           &	\textbf{60.98\%}  &	67.92\%          &	\textbf{50.81\%}    &	23.69\%           \\
Transformer   &	31.85\%             &	25.24\%       &	45.35\%           &	69.44\%          &	15.46\%             &	11.70\%            \\
\hline
\hline
CINO-base-v2  &    74.44\%          &-                & 75.04\%             &    69.44\%       &    72.82\%           &    73.08\%             \\
MiLMo-base   &	\textbf{91.62\%}    &-                &	\textbf{88.15\%}  &	\textbf{92.81\%} &	\textbf{82.05\%}    &	\textbf{73.34\% } \\
\hline
\end{tabular}
\end{center}\renewcommand{\tablename}{Table}
\caption{\label{font-table} Text classification results on WCM}
\end{table*}

\begin{figure*}[h] 
\setlength{\abovecaptionskip}{1pt}
\setlength{\belowcaptionskip}{2pt}
    \centering 
    \includegraphics[scale=0.8]{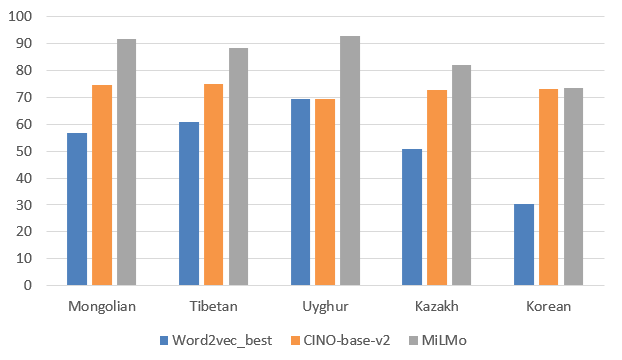}
    \renewcommand{\figurename}{Figure}
    \caption{Text classification results on WCM} 
    \label{wolf} 
\end{figure*}

From Table V, we can see that DPCNN has the best effect of 49.15$\%$ on Mongolian dataset. On Tibetan syllable level data, TextCNN has the best classification F1 of 39.60$\%$. On the Tibetan word level data, the effect of DPCNN reached the best 34.51$\%$. On Uyghur data, TextRNN has the best effect of 42.19$\%$. In Kazakh, the DPCNN model has the best effect of 30.13$\%$. TextCNN has the best effect of 53.17$\%$ in Korean. Overall, the DPCNN model achieves the best effect on various datasets, while Transformer performs poorly. The main reason is that the complex network structure of Transformer need more training data, while the number of classification datasets constructed in this paper is relatively small.

\subsection{Classification based on MiLMo}
In this paper, we use the trained XLM model for the downstream experiment of text classification. We first use MiLMo model to encode classification text, and obtain the representation vector $E={e_{1},e_{2},...,e_{256}}$, containing multilingual common information and text information. Then use the "linear + softmax" structure in the text classification layer for text classification, and the vector representation obtained in the coding layer is fed to the linear to obtain the vector $ s=Linear(E)$, and we use the softmax function to calculate the probability of text categories, $ p=softmax(s) $. After that cross entropy is used to calculate the loss function value of text classification $loss = ylogp$, where $y$ is the real result and $p$ is the prediction result.
This paper constructs the MiLMo model, and uses it to conduct experiments on multilingual classification datasets. The datasets are preprocessed by BPE. Then MiLMo model is used to encode the processed the data, and the "linear + softmax" structure is used to classify it. The experimental results are shown in Table VI, where word2vec\_Best is the most effective classification result of word2vec model in five minority languages.

From Table VI, we can see that the text classification based on the pre-trained model has the highest F1 of 85.98$\%$ in Korean and 71.34$\%$ in Kazakh. The text classification effect in five languages is higher than that of word2vec.

To further explore the effectiveness of the multilingual pre-trained model proposed in this paper this paper extracts Mongolian, Tibetan, Uyghur, Kazakh and Korean from the WCM for text classification experiments, and uses CINO-base-v2, word2vec and MiLMo model for comparison. CINO-base-v2 model is a multilingual pre-trained model for ethnic minorities released by HIT·iFLYEK Language Cognitive Computing Lab. The final experimental results are shown in Table VII and Figure II.

From Table VII and Figure II, we can see that our model MiLMo model has achieved the best results in five datasets. The classification effect of CINO-base-v2 in Tibetan, Korean, Mongolian and Kazakh is higher than that of word2vec, but lower than that of MiLMo model. In Uyghur, due to the unbalance distribution of WCM, the article category is mainly focused on the "geography". The classification F1 of CINO-base-v2 and word2vec is the same as 69.44$\%$, while the classification F1 of MiLMo model is 92.81$\%$, which shows that our model can still achieve good results on small-scale datasets. At present, the MiLMo model architecture trained in this paper only includes 12 layers of Transformer block. In future research, we will further release the MiLMo-large model based on 24 layers of Transformer block, which will further improve the effect on downstream tasks.

\section{Conclusion}
The multilingual pre-trained model provides support for various languages with rich resources. However, due to the rich morphology of minority languages, different grammatical rules, and difficult data acquisition, various natural language processing tasks for minority languages are still at the initial stage. To solve the above problems, this paper takes Mongolian, Tibetan, Uyghur, Kazakh, and Korean as examples. We obtain relevant data from relevant books, relevant documents of the NPC and CPPCC sessions and government work reports, and construct a multilingual dataset after data cleaning, and construct a multilingual pre-trained model MiLMo for ethnic minorities. To verify the effectiveness of the pre-trained model, this paper trains word2vec on five minority languages, and uses the word2vec word representation and the pre-trained model for text classification experiments. The experimental results show that the classification effect of the pre-trained model on five languages is better than word2vec.

\end{document}